\documentclass[10pt, a4paper]{article}

\usepackage{lrec-coling2024} 

\title{Testing MediaPipe Holistic for Linguistic Analysis \\ of Nonmanual Markers in Sign Languages}

\name{Anna Kuznetsova\textsuperscript{1}, Vadim Kimmelman\textsuperscript{2}} 

\address{\textsuperscript{1}University of Trento, \textsuperscript{2}University of Bergen \\
         \textsuperscript{1}Amsterdam, Netherlands, \textsuperscript{2}Bergen, Norway \\
         kuzannagood@gmail.com, vadim.kimmelman@uib.no\\}

\abstract{
Advances in Deep Learning have made possible reliable landmark tracking of human bodies and faces that can be used for a variety of tasks. We test a recent Computer Vision solution, MediaPipe Holistic (MPH), to find out if its tracking of the facial features is reliable enough for a linguistic analysis of data from sign languages, and compare it to an older solution (OpenFace, OF). We use an existing data set of sentences in Kazakh-Russian Sign Language and a newly created small data set of videos with head tilts and eyebrow movements. We find that MPH  does not perform well enough for linguistic analysis of eyebrow movement -- but in a different way from OF, which is also performing poorly without correction. We reiterate a previous proposal to train additional correction models to overcome these limitations. 
 \\ \newline  }

\begin{document}

\maketitleabstract
\section{Introduction}

Recent advances in Deep Learning have substantially improved the quality of Computer Vision (CV) solutions, making possible reliable tracking of human bodies and faces in video recordings and video stream. One popular CV package, OpenPose \citep{cao2018openpose}, has been used in a large variety of applications including Automatic Sign Language Recognition, Translation and Generation (SLRTG) \citep{adeyanju_machine_2021}. Another package, OpenFace \citep{baltrusaitis_openface_2018, zadeh2017convolutional}, was created as an extension of OpenPose with the task of emotion recognition; it has the additional functionality of tracking the head rotation (along the three axes) and 3D reconstruction of the head model from 2D video data. More recently, another solution has been published, namely MediaPipe \citep{lugaresi_mediapipe_2019}, including MediaPipe Holistic which tracks the body, the hands, and a large number of facial landmarks in 2D video data. 

It is clear that CV solutions can be used for research purposes, including in linguistics, and especially in the fields of gesture studies and sign language linguistics \citep{borstell_extracting_2023,trujillo_toward_2019}. CV solutions have been used most frequently and successfully for the tasks of SLRTG \citep{morgan_facilitating_2022}. 

Additionally, since it is now possible to automatically track the body, hands, and faces of signers and gesturers, researchers can use these tools to perform measurements of the motion and position of the subjects in video recordings \citep{borstell_extracting_2023,trujillo_toward_2019}. This has the potential to solve several important practical problems: 1) Automatic tracking is much less time consuming than manual annotation and 2) Automatic tracking can be more reliable than manual annotation. However, this potential is somewhat hypothetical. It is not clear whether the CV solutions are reliable/precise enough to enable linguistic analysis. 

This is especially pertinent for the issue of analyzing facial expressions which play an important role in sign language production and comprehension \citep{pfau_nonmanuals:_2010} but are produced by small articulators (e.g. eyebrows) which can be difficult to track precisely \citep{Kimmelman2020}.

In this paper, we specifically test MediaPipe Holistic (MPH) and compare it with OpenFace (OF) for the task of tracking eyebrow position in sign language recordings. Our previous research \citep{kuznetsova.etal2021Using, kuznetsova.etal2022Functional} has shown that OF can be used to analyze eyebrow position and head tilt used to mark questions in a sign language; however, it was only possible to use OF after training an additional correction model to counteract distortions introduced by head tilts. In this study, we use the same data set to see if MPH, being a more recent solution developed by a large company, still suffers from the same distortions. We also record an additional small data set to specifically test the effects of head tilts on eyebrow position estimation.

\section{Methods} 
\label{sec:methods}

We analyzed two data sets: a data set of utterances in Kazakh-Russian Sign Language (KRSL) produced by 9 signers \citep{kuznetsova.etal2021Using}, which is a part of the K-RSL dataset \cite{imashev-etal-2020-dataset}, and a newly recorded data set of head tilts and eyebrow raises produced by a single subject, both analyzed with OF and MPH. 

\subsection{Data Set 1: KRSL}

The KRSL data set contains video recordings of 9 signers of KRSL producing 10 sentences in three conditions (as a statement, a polar question, and a content question). In a previous study \citep{kuznetsova.etal2021Using}, we analyzed this data set with OpenFace and found out that polar questions are accompanied by eyebrow raise on the whole sentence and two downward head movements, content questions are marked with backward head tilt and eyebrow raise on the question sign, and statements are unmarked. This is similar to how questions are marked in some other sign languages, see \citet{cecchetto_sentence_2012}. 

\begin{figure*}[!ht]
    \centering
    \includegraphics[width=\textwidth]{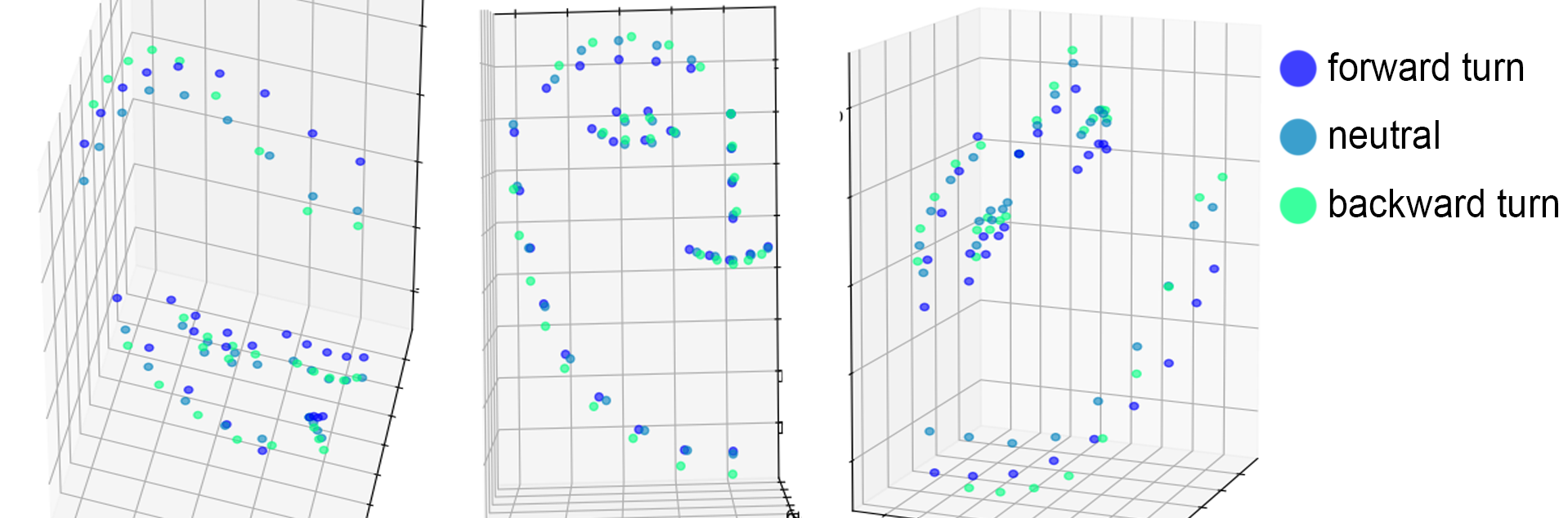}
    \caption{Head models based on OF outputs demonstrating distortion due to head pitch.}
    \label{fig:OFheads}
\end{figure*}

Another important finding, however, was that OF 3D reconstruction does not actually work for the purposes of linguistic analysis of the data directly. Specifically, we observed that the eyebrow position relative to the eye was strongly affected by the head tilt (Figure \ref{fig:OFheads}). This led to a distortion of the eyebrow position assessment which almost fully obfuscated the pattern described above.

We proposed a solution which, briefly, consisted of training a model using manually selected sentences without eyebrow movement which would predict eyebrow position based on head rotation. This model thus predicts the distortion of the eyebrow position due to head tilts. When applied to sentences with eyebrow movement this model corrects for this distortion. The resulting assessment of the eyebrow position is in agreement with the actual data. Figure \ref{fig:KRSL} in Section \ref{sec:results} shows the corrected and uncorrected outputs of OF. 

For the current study, we simply applied MPH to the same data set and analyzed the outputs for the same sentence types. Since MPH's face model has a much larger number of landmarks, a direct comparison is not possible. However, it is possible to compare the results using either single landmarks or by averaging over multiple landmarks in the MPH outputs.

\subsection{Data Set 2: Head and Eyebrow Movements}

As we discuss below, the MPH analysis of the KRSL data set looked promising as it showed the differences between sentence types in general agreement with the corrected results from OF. This indicated that MPH did not suffer from distortions in the context of head tilts. We decided to further test this directly by recording videos with and without eyebrow raise and different head movements. 

Specifically, we recorded a single person producing head pitches (up and down) with raised and with neutral eyebrow position.\footnote{We also recorded head yaws and rolls with neutral eyebrow position, and head pitches with variable eyebrow position. However, we do not further analyse these files in this report.} We applied MPH (to assess eyebrow location) and OF (to assess eyebrow location and head pitch) to this data set. 

\subsection{Data Preparation and Analysis}

In order to assess the effect of head tilts on eyebrow raise, we used the following methods. For the KRSL data set, we simply reconstructed the eyebrow movement patterns to replicate the findings from the previous study and graphically compared the results. The eyebrow distance was calculated using inner and outer eyebrow points and two inner points of the eyes -- we drew a line through the eye points and calculated the distance from eyebrow points to this line.

For the new data set, first, we calculated the standard deviation of the mean eyebrow distance (separately for inner and outer eyebrow landmarks) by comparing each measure of distance to the distance in the initial frame of the recording with no tilt. If the solution works perfectly, then the distance with and without tilts should be the same, and the deviation should be minimal. For comparison of different camera distances (close, middle, far) we scaled the eyebrow distances and rotation angles between 0 and 1 in each camera distance group. To test whether the differences are statistically significant, we conducted t-tests to compare the deviation in each video (separately for inner and outer eyebrow points) to zero, and to compare the OF and MPH deviations to each other.

Second, we explored the data graphically in two ways. (1) We created visualizations of the distortions introduced by different types of head tilts by visualizing the model of the eyebrows, eyes and nose in 3D space at three different time points (first frame, middle of the head pitch, peak of the head pitch). The face models were rotated to the neutral position by reverting the head rotation angles and centred around the upper nose point. (2) We created visualizations of the scaled eyebrow distance measures for the different video recordings in order to compare the videos with and without eyebrow raise in the context of the pitches upward and downward. 

Data preparation, analysis and visualization were conducted in Python and R \citep{r_core_team_r_2022,wickham_welcome_2019}. The video and data files and scripts to reproduce the results, as well as the full set of visualizations, can be found \href{https://osf.io/jvc42/}{here}.

\section{Results}
\label{sec:results}

\subsection{KRSL} 

\begin{figure*}[!ht]
    \centering
    \includegraphics[width=\textwidth]{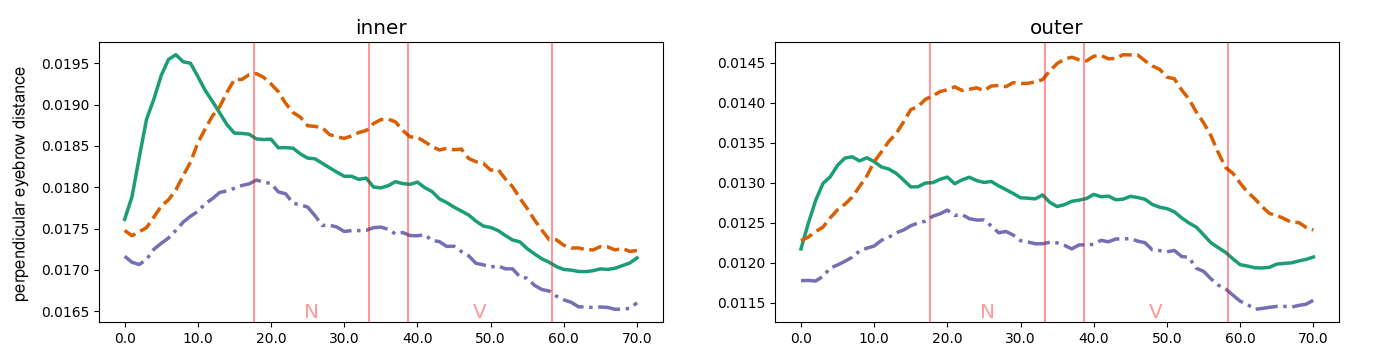}
    \includegraphics[width=\textwidth]{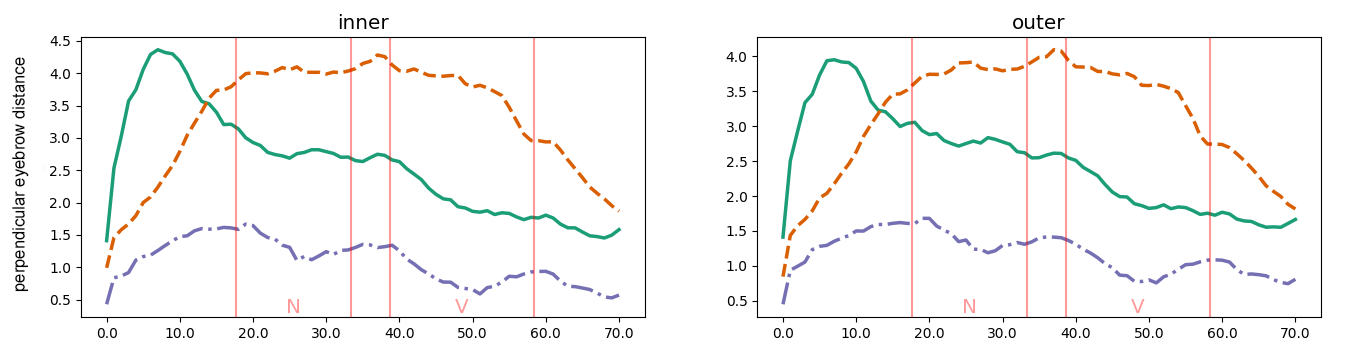}
    \includegraphics[width=\textwidth]{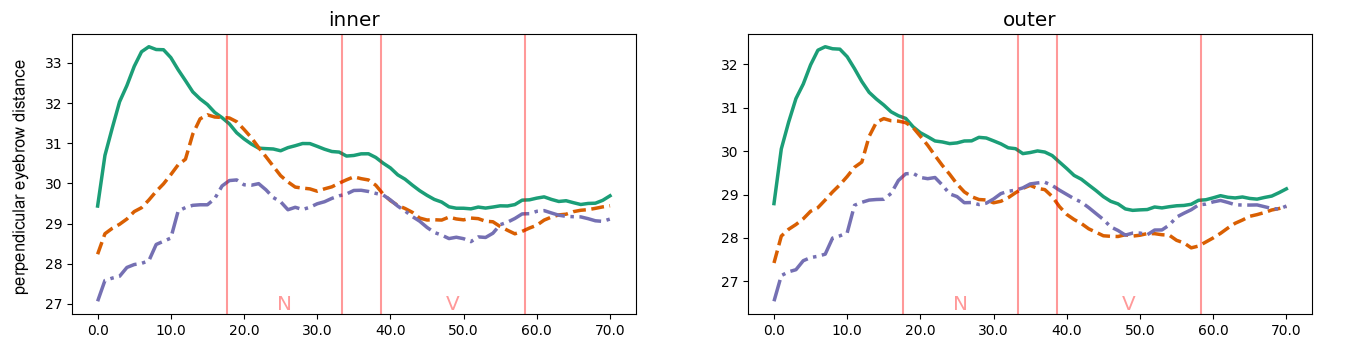}
    \caption{Eyebrow marking for different types of sentences in the KRSL data set. Top panel: MPH. Middle panel: OF corrected. Lower panel: OF non-corrected. Colors represent sentence types (orange: polar question, green: content question, purple: statement). N: noun, V: verb. Left column: inner eyebrow distance, right column: outer eyebrow distance. X-axis is frame number normalized to 70.}
    \label{fig:KRSL}
\end{figure*}

The results of applying MPH to the KRSL data set can be seen in Figure \ref{fig:KRSL}. The top panel shows the results of MPH application to the data. It is clear that it more closely resembles the results of OF with a correction model (middle panel) than OF without correction (lower panel). Specifically, the MPH outputs also clearly show eyebrow raise on the question sign in content questions, and eyebrow raise on the whole sentence in polar questions. Note however some issues: (1) the differences between sentence types are diminished in comparison to the middle panel, (2) there is a strong difference between inner and outer eyebrow distances. 

Despite the differences between MPH and OF corrected, since the overall pattern is similar, the application of MPH looks promising. If it were possible to use MPH to directly measure eyebrow position, also in the presence of head tilts, without additional correction, this would be a very practical solution for sign language linguistics. However, we decided to further test whether head tilts introduce distortion in eyebrow position estimation directly in the new data set. 

\subsection{Head and Eyebrow Movements} 

\subsubsection{Deviations Due to Head Movement}

As described in Section \ref{sec:methods}, we used video recordings of a single subject performing head movements with and without eyebrow raise, filmed from three different distances. We analyzed these videos with OF to estimate eyebrow position and head tilt, and with MPH to estimate eyebrow position.

We analyzed the estimated eyebrow distance in videos without eyebrow raise by comparing the eyebrow distance throughout the video with the eyebrow distance in the initial part which contained no head rotation. We calculated average deviation for inner eyebrow points and outer eyebrow points in MPH and OF, see the results in Table \ref{table:deviations}.

\begin{table*}[!ht]
    \centering
    \begin{tabular}{lcccc}
         \textbf{Type}&  \textbf{MPH inner}&  \textbf{OF inner}&  \textbf{MPH outer}& \textbf{OF outer}\\
         close pitch down&  \textcolor{red}{0.103}&  0.079&  \textcolor{red}{0.19}& 0.066\\
         close pitch up &  0.117&  \textcolor{red}{0.224}&  0.097& \textcolor{red}{0.265}\\
         far pitch down&  \textcolor{red}{0.1}&  0.053&  \textcolor{red}{0.098}& 0.054\\
         far pitch up &  0.109&  \textcolor{red}{0.2}&  0.08& \textcolor{red}{0.209}\\
         middle pitch down&  \textcolor{red}{0.097}&  0.062&  \textcolor{red}{0.161}& 0.054\\
         middle pitch up&  0.096&  \textcolor{red}{0.211}&  0.064& \textcolor{red}{0.26}\\
    \end{tabular}
    \caption{Deviation of eyebrow distance, no eyebrow raise, with different head pitches at different distances, with OF and MPH. The higher deviation between in pairwise comparison of MPH to OF is in red. The range of possible deviations is (0,1).}
    \label{table:deviations}
\end{table*}

We have conducted t-tests comparing the deviations in each case to each other (between OF and MPH), and to zero. In each case, the difference is highly statistically significant. Please see the code in the repository linked above. 

This table shows several puzzling patterns. First, it is clear that MPH is still susceptible to bias in eyebrow position estimation in the presence of head tilts. Second, while MPH performs considerably better than OF for the upward head pitch, OF actually outperforms MPH for downward head pitch. Finally, MPH performs worse for outer eyebrow estimation than for inner eyebrow estimation, which is not the case for OF. 

We also analyzed the videos with eyebrows raised the same way as the videos with no eyebrow raise. Note that in these videos the eyebrows are raised throughout and there is no eyebrow movement, while the head is moving. The results are reported in Table \ref{table:deviationsER}. While there are still significant differences between MPH and OF (and all the deviations are significantly different from zero), the differences are not the same as in Table \ref{table:deviations}.

\begin{table*}[!ht]
    \centering
    \begin{tabular}{lcccc}
         \textbf{Type}&  \textbf{MPH inner}&  \textbf{OF inner}&  \textbf{MPH outer}& \textbf{OF outer}\\
         close pitch down&  \textcolor{red}{0.146}&  0.079&  \textcolor{red}{0.211}& 0.178\\
         close pitch up &  \textcolor{red}{0.322}&  0.24&  \textcolor{red}{0.374}& 0.261\\
         far pitch down&  \textcolor{red}{0.131}&  0.074&  \textcolor{red}{0.138}& 0.102\\
         far pitch up &  \textcolor{red}{0.268}&  0.096&  \textcolor{red}{0.201}& 0.097\\
         middle pitch down&  \textcolor{red}{0.118}&  0.098&  0.097& \textcolor{red}{0.13}\\
         middle pitch up&  \textcolor{red}{0.273}&  0.128&  \textcolor{red}{0.244}& 0.145\\
    \end{tabular}
    \caption{Deviation of eyebrow distance, with eyebrow raise, with different head pitches at different distances, with OF and MPH. The higher deviation between in pairwise comparison of MPH to OF is in red. The range of possible deviations is (0,1).}
    \label{table:deviationsER}
\end{table*}

Here, in all but one case (middle pitch down), the deviations for MPH are much higher than for OF, so it performs much worse than OF. Note that it is very surprising to find that the results are different when using the videos with or without eyebrow movement. It is also surprising to see MPH perform so badly in this condition. In the next section, we confirm this pattern by visual inspection.

\subsubsection{Visualizing the Deviations}

\begin{figure*}[!ht]
    \centering
    \includegraphics[width=\textwidth]{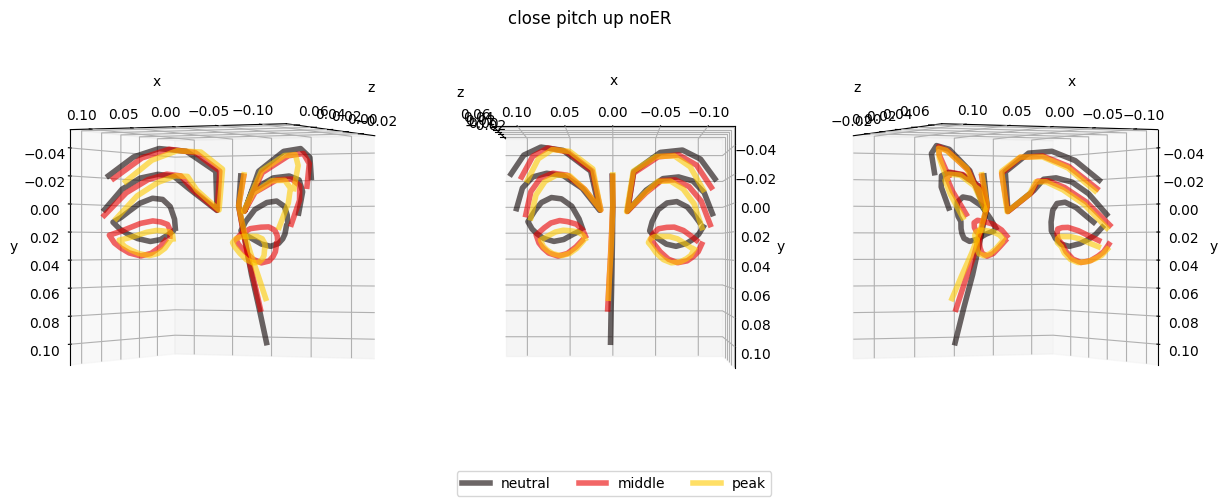}    
    \includegraphics[width=\textwidth]{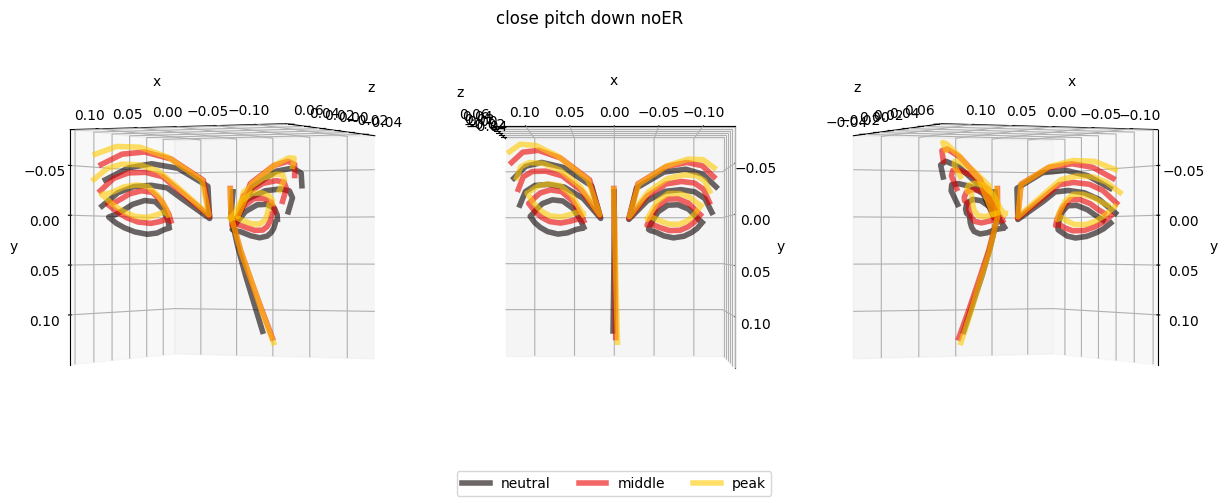}
    \caption{Head models based on MPH output, for close pitch up and close pitch down videos. Grey line: no pitch; red line: middle of pitch motion; yellow line: maximal pitch.}
    \label{fig:MPHheads}
\end{figure*}

Another way to see that MPH is susceptible to distortion due to head rotation is to look at visualizations of the head model (only eyebrows, eyes and nose landmarks included) produced from MPH outputs, see Figure \ref{fig:MPHheads}. This figure shows the head visualized in the beginning of a video (no pitch), in the middle of head movement, and in the peak of the head movement (upper panel: head pitch up, lower panel: head pitch down). 

It can be clearly seen that, with upward pitch, the head model is squished vertically, so the estimation of eyebrow distance is lowered. With downward pitch, the head model is stretched vertically, and the estimation of eyebrow distance is raised. This is the opposite pattern of distortion from OF (see Figure \ref{fig:OFheads} above), but it is still a very clearly present distortion.

\begin{figure*}[!ht]
    \centering
    \includegraphics[width=0.7\textwidth]{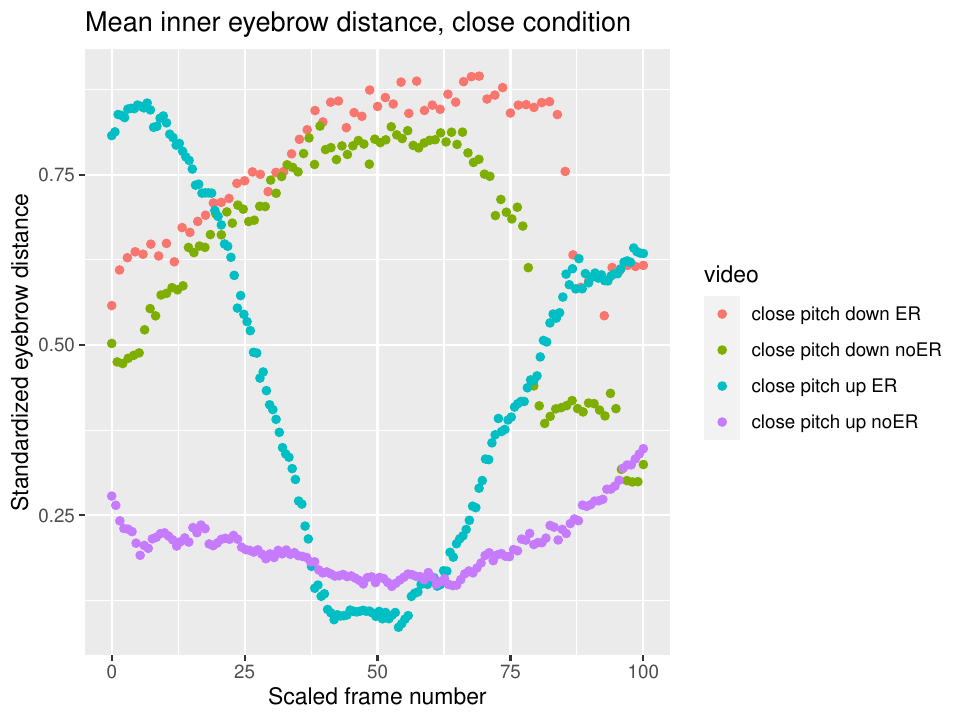}
    \caption{MPH eyebrow distance estimation for inner eyebrows in the close condition; head pitch up or down; with or without raised eyebrows. The distance and frame numbers are standardized.}
    \label{fig:innerClose}
\end{figure*}

Finally, yet another way to see the distortion is to visualize (standardized) eyebrow distance estimation for videos with and without eyebrow movement and with upward or downward pitch. Consider Figure \ref{fig:innerClose} which shows inner eyebrow distance in the close condition. An interesting pattern emerges: with downward pitch, both raised and non-raised eyebrows are distorted approximately to the same extent. With upward pitch, however, there is a very strong distortion of raised eyebrows, but the distortion for non-raised eyebrows is much smaller. 

\begin{figure*}[!ht]
    \centering
    \includegraphics[width=0.7\textwidth]{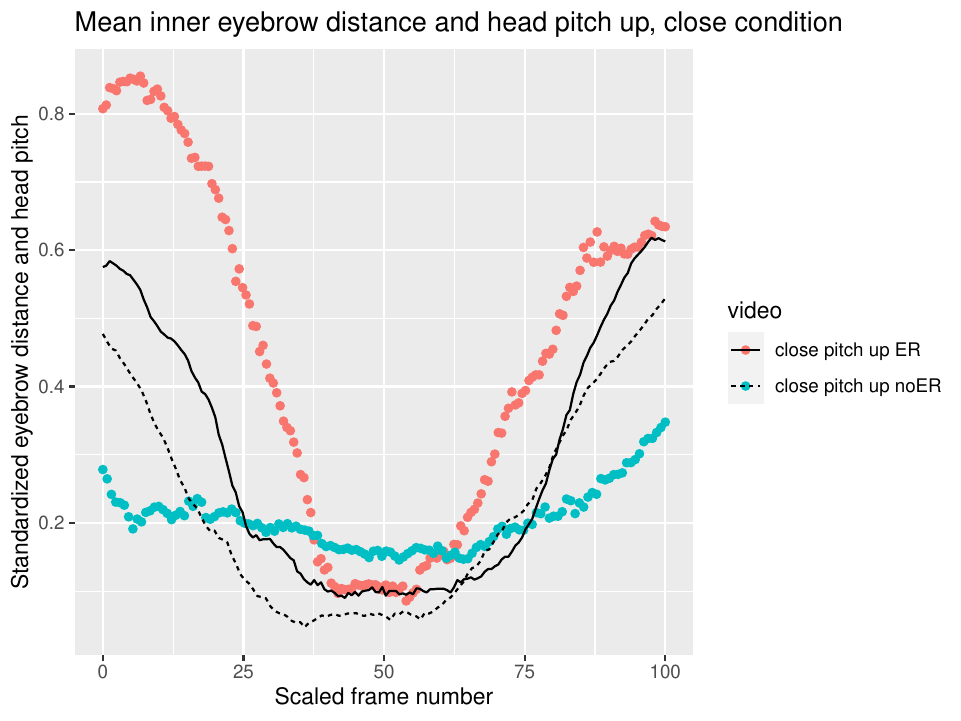}
    \caption{MPH eyebrow distance (colored dots) and head pitch (black lines) estimation for inner eyebrows in the close condition; head pitch up with or without raised eyebrows. The distances and frame numbers are standardized.}
    \label{fig:innerCloseUp}
\end{figure*}

To see that this pattern is not due to differences in head pitch itself, consider Figure \ref{fig:innerCloseUp} which shows eyebrow distance (as dots) and head pitch (as lines) in the videos with head pitch up with and without eyebrow movement. It can seen that, in both videos, the pitch is similar, but it affects the eyebrow distance estimation differently. The same pattern (the difference between videos with and without eyebrow raise for head pitch up) can be seen also at middle and far distances, although the pattern becomes less pronounced.

\begin{figure*}[!ht]
    \centering
    \includegraphics[width=0.7\textwidth]{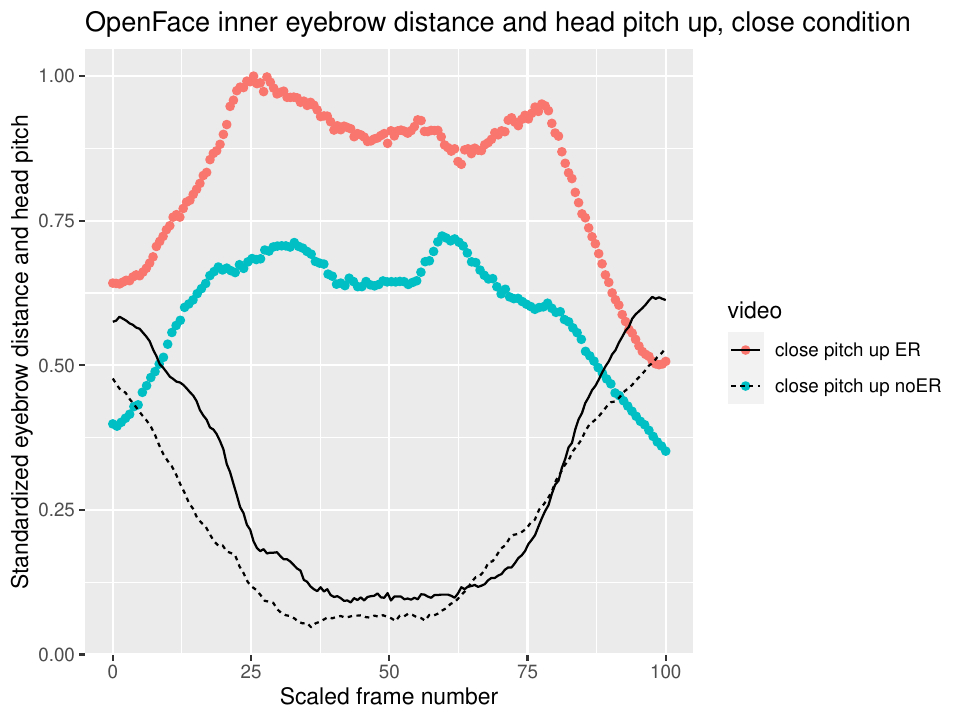}
    \caption{OF eyebrow distance (colored dots) and head pitch (black lines) estimation for inner eyebrows in the close condition; head pitch up with or without raised eyebrows. The distances and frame numbers are standardized.}
    \label{fig:OFCloseUp}
\end{figure*}

Compare also to Figure \ref{fig:OFCloseUp} showing the OF estimation of eyebrow distance in the same two videos. The pattern here is completely different. First, there is no difference between the videos with and without eyebrow raise: both are distorted to the same extent. Second, the distortion of the eyebrow distance is in the opposite direction from MPH, as mentioned above.

We conclude that both calculating deviations and graphical exploration demonstrate that MPH outputs and the measure of eyebrow distance that we derive from them are substantially distorted in the presence of head pitch. The distortion is complex and different from OF, but it is still very clearly present. 

\begin{figure*}[!ht]
    \centering
    \includegraphics[width=0.7\textwidth]{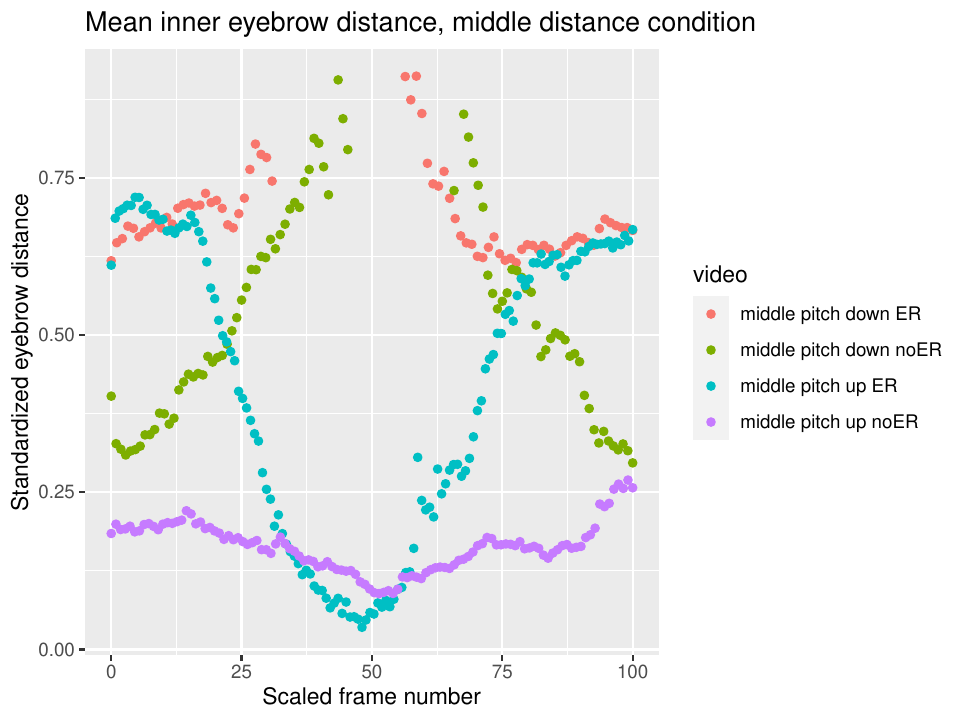}
    \caption{MPH eyebrow distance estimation for inner eyebrows in the middle condition; head pitch up or down; with or without raised eyebrows. The distance and frame numbers are standardized.}
    \label{fig:middle}
\end{figure*}

Finally, when graphically exploring the videos at middle and far distance, we also observed that MPH loses many data points with extreme upward head pitch (see Figure \ref{fig:middle}). This means that MPH does not reliably track the face in the presence of such head movements. This is especially relevant given that most sign language recordings are at a distance comparable to the far case in our study. 

\section{Discussion}

The main question of this study is whether MPH can track facial landmarks reliably enough and without distortions in the reconstructed 3D model so that its outputs can be used for linguistic analysis of nonmanual markers in sign languages. The answer to this question is a clear ``no''. 

We find clear evidence that the 3D model of the head that MPH outputs is distorted in the presence of head movements (specifically, head pitches). The distortion is so strong that it completely obscures very pronounced eyebrow position differences. In Figure \ref{fig:innerCloseUp}, we can see for example that, with a head pitch up, raised eyebrows at the peak of the pitch can be perceived as being at the same distance as the non-raised eyebrows. 

What is even more troubling is that the distortions are not describable by a simple trend. The distortion introduced by the pitch up movement is clearly stronger than the one introduced by pitch down. The distortion is also stronger for outer than inner eyebrows. Finally, the distortion of the pitch up movement is for some reason much stronger for raised than for non-raised eyebrows. In this respect, MPH works even worse than uncorrected OF because the distortions are more complex and thus might be more difficult to correct. 

With this result, a puzzle remains: why does the analysis of the KRSL data set with MPH look better than the uncorrected OF output? We think that this is a coincidence based on the co-occurrence pattern of the head and eyebrow movement in this specific data set. 

Specifically, for polar questions, the head pitches down while the eyebrows are raised, and for content questions, the head pitches up while the eyebrows are raised. For OF, the downward pitch leads to a decrease in eyebrow distance, and so polar questions with eyebrow raise become indistinguishable from statements without eyebrow raise. For MPH, the distortion goes in the opposite direction, as downward pitch leads to an increase in eyebrow distance, and so in the MPH output polar questions are different from statements. For pitch up in content questions, the effects are mirrored, and here we could expect MPH to obscure eyebrow raise, but because MPH is on average better in the presence of pitch up, the raise on the question sign is still visible (but mostly for outer eyebrows). In other words, while MPH happens to output a reasonable picture for the KRSL data set we analyzed, it is pure coincidence and does not indicate that MPH can be used directly for other data sets. 

Finally, we hope that the procedure of correcting the outputs that we proposed for OF in our previous work \citep{kuznetsova.etal2021Using, kuznetsova.etal2022Functional} can also be applied to the outputs of MPH. However, as discussed above, the distortions that MPH introduces are more complex and less linear than those in OF, and thus corrective models might not perform as well. We plan to further test this in future. 

\section{Bibliographical References}\label{sec:reference}

\bibliographystyle{lrec-coling2024-natbib}
\bibliography{lrec-coling2024-example}

\end{document}